\documentclass[twoside,onecolumn]{article}

\usepackage{tabulary,graphicx,times,caption,fancyhdr,amsfonts,amssymb,amsbsy,latexsym,amsmath}
\usepackage[utf8]{inputenc}
\usepackage{url,multirow,morefloats,floatflt,cancel,tfrupee,textcomp,colortbl,xcolor,pifont}
\usepackage[nointegrals]{wasysym}
\usepackage{subfigure}

\makeatletter

\usepackage{ifxetex}
\ifxetex\else
  \usepackage{dblfloatfix}
\fi

\@ifundefined{subparagraph}{
\def\subparagraph{\@startsection{paragraph}{5}{2\parindent}{0ex plus 0.1ex minus 0.1ex}%
{0ex}{\normalfont\small\itshape}}%
}{}

\def\URL#1#2{\@ifundefined{href}{#2}{\href{#1}{#2}}}

\def\UrlOrds{\do\*\do\-\do\~\do\'\do\"\do\-}%
\g@addto@macro{\UrlBreaks}{\UrlOrds}

\makeatother

\usepackage[paperheight=11in,paperwidth=8.3in,margin=2.5cm,headsep=.7cm,top=2.5cm]{geometry}
\usepackage[T1]{fontenc}

\renewenvironment{abstract}
	{\trivlist\item[]\leftskip0pt\par\vskip4pt\noindent
  	\textbf{\abstractname}\mbox{\null}\\}
	{\par\noindent\endtrivlist}

\def\keywords#1{\par\medskip\par\noindent\textbf{Keywords}: #1\par}

 \date{} \emergencystretch 8pt

\makeatletter
\def\author#1{\gdef\@author{\hskip-\tabcolsep%
	\parbox{\textwidth}{\raggedright\bfseries#1\\[1pc]}}}
\def\address[#1]#2{\g@addto@macro\@author{\\\hskip-\tabcolsep\parbox{\textwidth}{\raggedright%
	\normalsize\normalfont\textsuperscript{#1}#2}}}
\let\addresslink\textsuperscript
\def\correspondence#1{\g@addto@macro\@author{\\\hskip-\tabcolsep\parbox{\textwidth}{\raggedright%
	\vspace*{10pt}\normalsize\normalfont~\\#1~\\[12pt]}}}
\def\email#1{\g@addto@macro\@author{\\\hskip-\tabcolsep\parbox{\textwidth}{\raggedright%
	\normalsize\normalfont Emails: #1}}}

\def\title#1{\gdef\@title{\vspace*{-30pt}%
	\raggedright\textbf{\@journaltitle}~\\%
  \raggedright\bfseries\ifx\@articleType\@empty\vspace*{20pt}\else%
  \vspace*{20pt}\@articleType\vspace*{20pt}\\\fi#1}}
\let\@journaltitle\@empty \def\journaltitle#1{\gdef\@journaltitle{{\normalfont\itshape#1}}}
\let\@articleType\@empty \def\articletype#1{\gdef\@articleType{{\normalfont\itshape#1}}}

\let\@runningHead\@empty \def\RunningHead#1{\gdef\@runningHead{{\normalfont #1}}}

\usepackage{fancyhdr}
\fancypagestyle{headings}{\fancyhf{}
  \fancyhead[R]{\itshape\@runningHead}
  \fancyfoot[C]{\thepage}}
\makeatother

\usepackage[%
	numbers,sort&compress%
  ]{natbib}

\usepackage{float,xcolor}

\articletype{Research Article} 

\begin{document}

\newcommand{\blue}[1]{{\leavevmode\color{blue}[#1]}}
\newcommand{\green}[1]{{\leavevmode\color{green}[#1]}}
\newcommand{\olive}[1]{{\leavevmode\color{olive}#1}}
\newcommand{\xx}{\boldsymbol{x}}
\newcommand{\yy}{\boldsymbol{y}}
\newcommand{\zz}{\boldsymbol{z}}

\title{
Enhancing Fingerprint Image Synthesis with GANs, Diffusion Models, and Style Transfer Techniques
}

\author{%
	W. Tang\addresslink{1,}\footnote{Both first authors contributed equally},\;
  	D. Figueroa\addresslink{1,*}, 
   D. Liu\addresslink{1},
   K. Johnsson\addresslink{2} and
  	A. Sopasakis\addresslink{1}
    }
		
\address[1]{Mathematics, Lund University}
\address[2]{Precise Biometrics AB}

\correspondence{Correspondence should be addressed to 
    	A. Sopasakis: alexandros.sopasakis@math.lth.se}


\email{macro.wz.tang@gmail.com (W. Tang), figueroall.diego@gmail.com (D. Figueroa), 
donglin.liu@math.lth.se (D. Liu),
kerstin.johnsson@precisebiometrics.com (K.Johnsson),
alexandros.sopasakis@math.lth.se (A. Sopasakis)}%

\RunningHead{
Fingerprint Synthesis from Diffusion Models and GANs
}

\maketitle 

\begin{abstract}
We present novel approaches involving generative adversarial networks and diffusion models in order to synthesize high quality, live and spoof fingerprint images while preserving features such as uniqueness and diversity. We generate live fingerprints from noise with a variety of methods, and we use image translation techniques to translate live fingerprint images to spoof. To generate different types of spoof images based on limited training data we incorporate style transfer techniques through a cycle autoencoder equipped with a Wasserstein metric along with Gradient Penalty (CycleWGAN-GP) in order to avoid mode collapse and instability. 
We find that when the spoof training data includes distinct spoof characteristics, it leads to improved live-to-spoof translation.

We assess the diversity and realism of the generated live fingerprint images mainly through the Fr\'{e}chet Inception Distance (FID) and the False Acceptance Rate (FAR). Our best diffusion model achieved a FID of 15.78. The comparable WGAN-GP model achieved slightly higher FID while performing better in the uniqueness assessment due to a slightly lower FAR when matched against the training data, indicating better creativity. Moreover, we give example images showing that a DDPM model clearly can generate realistic fingerprint images.


\end{abstract}

\section{Introduction\label{sec:intro}}


Biometric technologies have gained widespread adoption across various fields for identity authentication purposes. Fingerprint, facial, and iris recognition based on biometric features are increasingly recognized as secure, confidential, and convenient alternatives to traditional identification methods. This is due to the inherent characteristics of biometric features, which are generally measurable, verifiable, and unique. However, the collection of fingerprint data presents challenges in terms of cost and time, as a significant number of high-quality fingerprint images are required to develop and optimize fingerprint algorithms. Additionally, the collection, utilization, and storage of biometric information raises complex issues regarding privacy and compliance with regulations such as the General Data Protection Regulation (GDPR) in Europe. These policies ensure the protection of individuals' privacy, yet make it more challenging to access and collect human characteristics for biometric authentication purposes.

``Live fingerprints'' and ``spoof fingerprints'' are terms related to biometric security, specifically in the context of fingerprint recognition systems. These terms are used to describe different types of fingerprints and the challenges associated with ensuring the authenticity of the fingerprint being presented for identification.
Live fingerprints refer to 
authentic fingerprints captured directly from a person's finger. In biometric systems, such as fingerprint scanners used for access control or authentication purposes, live fingerprints are the actual physical impressions of the ridges and valleys present on an individual's fingertip. These fingerprints are unique to each person and are used for accurate identification and verification.
Spoof fingerprints, on the other hand, are artificially created or replicated fingerprints that are designed to deceive fingerprint recognition systems. These could be crafted using various materials, such as gelatin, silicone, or even printed on a piece of paper \cite{spoof-handbook}. The goal of a spoof fingerprint is to mimic the characteristics of a live fingerprint in order to gain unauthorized access to a system, device, or facility.
Spoofing is a security concern in biometric systems, as it can undermine the effectiveness of fingerprint-based authentication. 

In order to alleviate the data collection problem, we introduce a method for the artificial synthesis of fingerprint patches using a state-of-the-art image generator, the Denoising Diffusion Probabilistic Model (DDPM) \cite{diffusion:DDPM}. DDPMs stand at the forefront of image generation, generating specific images through iterative denoising of pure noise samples. To provide a more comprehensive analysis of DDPMs and the generated results, we also construct and train a state-of-the-art Wasserstein Generative Adversarial Network with Gradient Penalty (WGAN-GP) for comparative purposes.

Additionally, we propose an alternative method for generating fingerprint patches from limited datasets by transforming live fingerprints into spoof fingerprints and vice versa. We leverage state-of-the-art style transfer techniques, such as cycleGAN and cycleWGAN-GP, to blend the global structure of one fingerprint with the local features of another. This process yields a fingerprint that is visually realistic, distinctive, and in line with the corresponding live fingerprint from which it originates.
Furthermore, using the proposed approach it should be possible to generate fingerprints under different environmental or other external conditions such as surface temperature, condensation, contaminants, and surface residues. 


In our evaluation of the synthetic fingerprints, we employ a range of well-established metrics, including the Fréchet Inception Distance (FID) 
\cite{metric:FID}, the Kernel Inception Distance (KID) 
\cite{metric:KID}, and the False Acceptance Rate (FAR) \cite{metric:FAR}. Additionally, we incorporate recently proposed metrics such as Precision, Recall, Density, and Coverage (PRDC) \cite{metric:pr}\cite{metric:dc}. These metrics serve as valuable complements to the existing assessment results. 
Another common metric, the Inception Score (IS) \cite{metric:IS}, is primarily used for evaluating images generated by models trained on the ImageNet dataset across 1000 different categories. However, IS's interpretability becomes ambiguous when it is applied to a dataset consisting of a single category. Consequently, we decided to not use the IS metric for the evaluation of fingerprint synthesis.



To carry out the analysis, we primarily utilize private datasets sourced from Precise Biometrics, a Swedish company. Due to confidentiality agreements and GDPR restrictions, we are not able to present real fingerprints in this article. 
Nevertheless, we will be able to present one set of synthetic fingerprint results generated from a private training dataset.


We begin in Section \ref{sec:sota} with an overview of the state of the art 
and how our contribution relates to previous work. We then describe the methodology of our approach in Section \ref{sec:methods} and present a number of synthetic fingerprints generated by our models in the Results in Section \ref{sec:results}. We discuss these results in Section \ref{sec:discussion} and provide our conclusions in Section \ref{sec:conslusions}.

\section{State of the Art\label{sec:sota}}

The field of fingerprint synthesis has experienced significant development since the early 21st century. In 2000, the groundbreaking work by R. Cappelli et al. \cite{related:2000} introduced a novel approach to generating fingerprints using Gabor-like space-variant filters that could be tuned according to the underlying ridge orientation. Building upon this, two years later, their work was further enhanced to derive a series of impressions from a master fingerprint, facilitating the practical construction of large-scale fingerprint databases \cite{related:2002}. Subsequently, in 2007, a modified Gabor filter was employed for ridge pattern generation \cite{related:2007}. Concurrently, researchers explored the synthesis of iris patterns, employing techniques such as principal component analysis (PCA) \cite{related:2004iris} and Markov Random Fields \cite{related:2005iris}. Additionally, numerous studies focused on fingerprint reconstruction based on minutiae templates, resulting in a substantial body of literature \cite{related:2007reconstruction, related:2007reconstruction2, related:2007reconstruction3, related:2009reconstruction, related:2010reconstruction, related:2014reconstruction}. In 2013, a seminal paper demonstrated the effectiveness of a method for modeling and generating synthetic fingerprint textures \cite{related:2013}.

Since the introduction of Generative Adversarial Networks (GANs) in 2014 \cite{GAN}, deep learning approaches have gained increasing popularity in the field of fingerprint synthesis. Finger-GAN \cite{related:2018fingerGAN} was an early endeavor that successfully implemented GANs to generate realistic fingerprints. This breakthrough led to the exploration of diverse GAN models for fingerprint synthesis. For instance, two-stage GANs \cite{related:2019two-stageGAN} and lightweight GANs \cite{related:2020lightGAN} were developed based on the original standard GANs \cite{GAN}. In a recent study, DCGAN-based approaches \cite{related:2021DCGAN} proposed two networks, namely Alex-GAN and VGG-GAN, which improved the performance of DCGAN by leveraging modified versions of convolutional neural networks, AlexNet \cite{related:alex} and VGG11 \cite{related:vgg}, to enhance the discriminator. Moreover, Striuk and Kondratenko \cite{related:2021ADCGAN} investigated Adaptive Deep Convolutional GAN (ADCGAN) for fingerprint synthesis. Another notable contribution was the progressive growth-based GANs utilized in the development of the Clarkson Fingerprint Generator (CFG) \cite{related:2021CFG}, enabling the generation of realistic fingerprints at a resolution of up to 512x512 pixels. Additionally, recent studies implemented cycleGAN \cite{cycleGAN} to generate high-fidelity fingerprints \cite{related:2020cycleGAN1, related:2022cycleGAN2}. FingerGAN \cite{related:2022fingerGAN} proposed an approach that embedded skip-connected DAE (SC-DAE) in GANs as a generator, while SpoofGAN \cite{related:2023spoofGAN} focused on generating spoof fingerprints.


Our study aims to specifically address the challenge of data scarcity in this field by adopting the strategy of generating fingerprint patches as opposed to complete fingerprints. Our research centers on the direct synthesis of individual patches as well as the transformation of one fingerprint into another. In the context of live fingerprint patch synthesis, we recognize the inherent uniqueness of fingerprint patterns, thereby prompting our exploration of the application of diffusion models and in particular DDPMs \cite{diffusion:DDPM}. 
To the best of our knowledge, no prior research has been found that applies these concepts to fingerprint synthesis, as described earlier. 
The progressive denoising process undertaken by the DDPM facilitates the refinement of a randomly initialized noise image over a span of 1,000 iterations. In each iterative step, the introduction of random Gaussian noise maintains an element of stochasticity in the resulting image. This iterative scheme empowers the creation of a diverse collection of synthetic fingerprints, each with its own unique characteristics. 

An inherent limitation frequently associated with Generative Adversarial Networks (GANs) \cite{GAN} is the challenge of mode collapse, which becomes particularly pronounced in scenarios involving imbalanced datasets. The asymmetry in data distribution holds true in our case, where live fingerprints typically outnumber spoof fingerprints. To tackle this, we build upon the foundation of CycleGAN \cite{cycleGAN}, which is a model renowned for its adeptness in unsupervised and unpaired image-to-image translation tasks. Elevating this approach, we also turn to CycleWGAN-GP \cite{cycleWGAN}, an enhanced iteration of CycleGAN that boasts heightened resilience against mode collapse-related issues.

This manuscript builds upon the approach developed in the thesis work of two of the current contributing authors, which was conducted under the guidance of Dr. Sopasakis, Dr. Johnsson and Dr. Liu \cite{thesis}.


\section{Methods\label{sec:methods}}

In this section we describe the main methods behind our approach in producing live and spoof fingerprint images which are able to maintain uniqueness and complexity. We begin by introducing the mathematical notation and present our methodology involving mechanisms such as DDPMs and CycleWGAN-GP.  We also present the metrics which will allow us to evaluate the overall quality of the resulting fingerprints. 

\subsection{DDPMs\label{subsec:DDPMs}}

\newcommand{\xO}{\mathbf{x}_0}
\newcommand{\xT}{\mathbf{x}_T}
\newcommand{\hx}{\hat{\mathbf{x}}}
\newcommand{\bI}{\mathbf{I}}
\newcommand{\beps}{\boldsymbol{\epsilon}}
\newcommand{\btheta}{{\boldsymbol{\theta}}}
\newcommand{\baralpha}{\bar{\alpha}}
\newcommand{\bz}{\mathbf{z}}
\newcommand{\bO}{\mathbf{0}}

\begin{figure*}[ht!]
    \centering
    \includegraphics[width=0.75\textwidth]{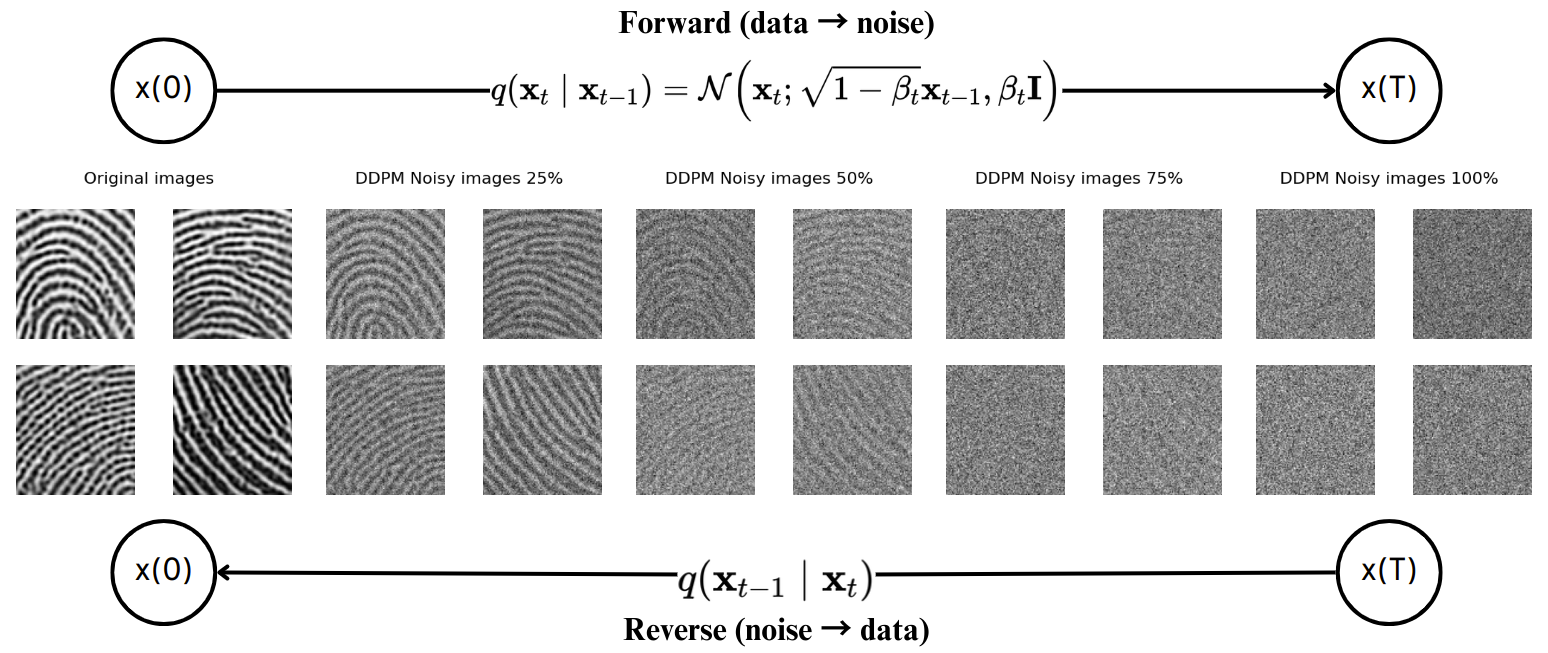}
    \caption{The forward and reverse processes involved in DDPMs. In the forward process, Gaussian noise is constantly added in each step. This becomes an isotropic Gaussian noise image in the end. In the reverse process, a random noise sample drawn from an isotropic Gaussian distribution is denoised at each level according to a deep learning model and, in theory, reconstituted to a clear image.}
    \label{fig:DDPM_process}
\end{figure*}

DDPMs are generative models that are based on the physical concept of diffusion where an initially clear image $\xO$ from a distribution of images $q(\xx)$, can be diffused into a pure noise image $\xT$ by iteratively adding noise, $\beps \sim \mathcal{N}(\bO, \bI)$, during $T$ time steps. Then, in the reverse process, noise is gradually removed from a noise image until a clear image is obtained. Image generation is formulated as sampling from posterior distributions through a neural network $q$. Figure \ref{fig:DDPM_process} presents examples of both the forward process (top row) as well as the reverse process (bottom row).

The denoising process is performed using a neural network $q$ with a U-net architecture.
The network is trained to estimate the noise part from images with varying levels of noise. The input to  network $q$ is the noisy image at a previous time step $t$, $\xx_t = \sqrt{\baralpha_t} \xO + \sqrt{1-\baralpha_t} \beps$, as well as the standard deviation $\beta_t$. The parameters for these equations are $\alpha_t = 1 - \beta_t$ and $\baralpha_t = \prod_{s=0}^t \alpha_t$ \cite{diffusion:DDPM, DDPMbeatGAN}. 
The output of the trained model is an estimate of the noise, $\beps_\btheta$, where $\btheta$ denotes the model weights. During this reverse process (also called sampling process), the estimated noise $\beps_\btheta$ is subtracted during each time step and additional noise $\bz \sim \mathcal{N}(\bO, \bI)$ is added as follows \cite{diffusion:DDPM, DDPMbeatGAN},
\begin{equation}
\xx_{t-1} = \frac{1}{\sqrt{\alpha_t}}\left( \xx_t - \frac{1 - \alpha_t}{\sqrt{1 - \baralpha_t}}\beps_\btheta(\xx_t, t) \right) + \sigma_t \bz,
\end{equation}
where $\sigma_t^2 = (1 - \alpha_t) (1 - \baralpha_{t-1}) / (1 - \baralpha_t)$ if $t > 0$ and $\sigma_t^2 = 0$ otherwise. The unknown hyperparameters $\beta_t$ which $\alpha_t$ depends upon are specified according to a predefined schedule \cite{diffusion:DDPM+}. In the experiments presented later we consider either a linear or a cosine schedule (see models vDDPM-v1 and DDPM-v2 in Table \ref{tab:models}). 

\subsection{WGAN-GP and CycleWGAN-GP\label{subsec:cycleWGAN}}


In this section we provide all the details about the loss functions behind each of the mechanisms we employ as well as the full objective function of the CycleWGAN-GP.

The traditional GAN loss is typically represented as a minimax game between a generator network, $G$, and a discriminator network $D$. The goal of the generator is to generate data that is indistinguishable from real data, while the discriminator aims to correctly distinguish between real and fake data. The loss function for the original GAN can be written as follows,
\begin{equation}
\mathcal{L}_{\text{GAN}}(G, D) = \mathbb{E}_{\xx \sim p_{\text{data}}(\xx)}[\log D(\xx)] + \mathbb{E}_{\zz \sim p_{\zz}(\zz)}[\log(1 - D(G(\zz)))],
\label{eq:origGAN}
\end{equation}
where $\xx$ represents real data samples drawn from the true data distribution $p_{\text{data}}(\xx)$, $\zz$ represents random noise samples drawn from a prior distribution $p_{\zz}(\zz)$ (usually Gaussian or uniform noise), $G(\zz)$ is the generated data obtained by passing noise $\zz$ through the generator $G$. Here $\log D(\xx)$ computes the log-probability that the discriminator assigns to real data and $\log(1 - D(G(\zz)))$ computes the log-probability that the discriminator assigns to fake data $G(\zz)$. The objective of the generator is to minimize the loss in (\ref{eq:origGAN}), while the objective of the discriminator is to maximize it.

In contrast, the Wasserstein GAN (WGAN) utilizes the Wasserstein distance \cite{Kantorovich-Rubinstein} as the objective function for training the generator and discriminator networks.
This distance metric quantifies the discrepancy between the probability distributions of real and fake data, offering greater stability and ease of optimization compared to the traditional GAN loss.
The WGAN loss is given by,
\begin{equation}
\mathcal{L}_{\text{WGAN}}(G, D) = \mathbb{E}_{\xx \sim p_{\text{data}}(\xx)}[D(\xx)] - \mathbb{E}_{\zz \sim p_{\zz}(\zz)}[D(G(\zz))].
\label{eq:WGANLoss}
\end{equation}
The generator tries to minimize this loss, while the discriminator tries to maximize it. The WGAN loss has several advantages, including better stability during training and improved convergence properties compared to the original GAN loss. It directly encourages the discriminator to provide meaningful and smooth scores for both real and fake data, enabling a more informative gradient signal for training.

It's important to note that in the WGAN framework, weight clipping or gradient penalty techniques are often used to enforce the Lipschitz constraint on the discriminator to ensure the Wasserstein distance is well-defined and the training remains stable.
WGAN-GP was first introduced in 2017 in \cite{WGAN-GP} where the authors proposed a modification to the loss function used in the original Wasserstein GAN (WGAN) \cite{WGAN},
\begin{equation}
\begin{aligned}
\mathcal{L}_{\text{WGAN-GP}}(G, D) =
\underbrace{
  \underset{\zz \sim p_{\zz}(\zz)}{\mathbb{E}}
  [D(G(\zz))] -
  \underset{\xx \sim p_{\text{data}}(\xx)}{\mathbb{E}}
  [D(\xx)]
}_\text{Original critic loss }  
+ \underbrace{
  \lambda_{\text{GP}} 
  \underset{\hat{\xx} \sim p_{\hat{\xx}}(\hat{\xx})}{\mathbb{E}}
  \left[ \left(\left\|\nabla_{\hat{\xx}} D(\hat{\xx})\right\|_2-1\right)^2 \right]
}_{\text {Gradient penalty }},
\end{aligned} 
\label{eq:WGAN-GPloss}
\end{equation}
where $\hat{\boldsymbol{x}}$ is a point sampled uniformly along straight lines between pairs of real and fake data samples $\hat{\boldsymbol{x}} = \epsilon \boldsymbol{x} + (1-\epsilon) G(\boldsymbol{z})$ where $\epsilon$ is drawn from a uniform distribution in the range $[0, 1]$, $\nabla_{\hat{\boldsymbol{x}}} D(\hat{\boldsymbol{x}})$ is the gradient of the discriminator $D$ with respect to $\hat{\xx}$ and $\lambda_{\mathrm{GP}}$ is a hyperparameter that controls the strength of the gradient penalty.

The CycleGAN (Cycle-Consistent Generative Adversarial Network) \cite{cycleGAN} introduces a cycle-consistency loss, $\mathcal{L}_{\text{cycle}}$,  in addition to the standard GAN losses in order to enable unpaired image-to-image translation. This loss enforces the idea that if an image is translated from domain $A$ to domain $B$ and then back to domain $A$, it should ideally be close to the original image. The overall loss function for the CycleGAN consists of several components,
\begin{equation}
\begin{aligned}
\mathcal{L}_{\text{CycleGAN}}(G_{A \to B}, G_{B \to A}, D_A, D_B) & =
\mathcal{L}_{\text{GAN}}(G_{A \to B}, D_B) + \mathcal{L}_{\text{GAN}}(G_{B \to A}, D_A) \\
& + \lambda_\text{cycle} \mathcal{L}_{\text{cycle}}(G_{A \to B}, G_{B \to A}),
\label{eq:new_cycleGAN}
\end{aligned}
\end{equation}
where $G_{A \to B}$ and $G_{B \to A}$ are the generators for translating images from domain $A$ to domain $B$ and vice versa, respectively, $D_A$ and $D_B$ are the discriminators for domains $A$ and $B$, respectively, $\lambda_\text{cycle}$ is a hyperparameter that controls the importance of the cycle-consistency loss relative to the adversarial losses. 
We note that the cycle-consistency loss $\mathcal{L}_{\text{cycle}}$ in (\ref{eq:new_cycleGAN}) is described as, 
$$
\mathcal{L}_{\text{cycle}}(G_{A \to B}, G_{B \to A}) 
= \mathbb{E}_{\xx \sim p_\text{data}(\xx)} \left[ \left\| \xx - G_{B \to A}(G_{A \to B}(\xx)) \right\|_1 \right]
+ \mathbb{E}_{\yy \sim p_\text{data}(\yy)} \left[ \left\| \yy - G_{B \to A}(G_{B \to A}(\yy)) \right\|_1 \right],
$$
based on sampling $\xx \sim p_\text{data}(\xx)$ from the data domain $A$ as well as sampling $\yy \sim p_\text{data}(\yy)$ from the data domain $B$. The cycle-consistency loss measures the difference between original images and images that have been translated and then translated back. This loss therefore helps ensure that the generated images are consistent with the original ones.
Finally, for certain translation tasks, it is also useful \cite{cycleGAN} to add the identity loss to the loss function, 
\begin{equation}
\begin{aligned}
\mathcal{L}_{\text {identity }}(G_{A \to B}, G_{B \to A}) = \mathbb{E}_{\yy \sim p_{\text {data }}(\yy)}\left[\|G_{A \to B}(\yy)-\yy \|_1\right]+\mathbb{E}_{\xx \sim p_{\text {data }}(\xx)}\left[\|G_{B \to A}(\xx)-\xx\|_1\right] . \label{eq:identity_loss}
\end{aligned}
\end{equation} 
This identity loss plays an important functional role for us. We use it to preserve the ridge structures of the fingerprints.

In an effort to improve the training stability of CycleGAN, \cite{cycleWGAN} introduces an objective function that incorporates the concepts outlined above.
This is mainly achieved by extending the concept of WGAN \cite{WGAN} and Improved WGAN \cite{WGAN-GP} while introducing gradient penalty terms \cite{WGAN-GP} based on ideas from WGAN as in (\ref{eq:WGAN-GPloss}). The full objective function of CycleWGAN-GP is therefore expressed as,

\begin{equation}
\begin{aligned}
\label{loss_cycleWGAN-GP}
\mathcal{L}_\text{CycleWGAN-GP} & \left(G_{A \to B}, G_{B \to A}, D_A, D_B \right)  \\
& =\mathcal{L}_{\text{WGAN-GP}}\left( G_{A \to B}, D_A \right) 
  +\mathcal{L}_{\text{WGAN-GP}}\left( G_{B \to A}, D_B \right) \\
 & + \lambda_\text{cycle} \mathcal{L}_{\text{cycle}}(G_{A \to B}, G_{B \to A}) 
 + \lambda_\text{identity} \mathcal{L}_\text{identity}(G_{A \to B}, G_{B \to A}),
\end{aligned}
\end{equation}
where the parameter $\lambda_\text{identity}$ controls the strength of the identity loss (\ref{eq:identity_loss}).


\subsection{Models and Datasets\label{subsec:MandD}}
For artificial synthesis of fingerprint patches we created, trained and tested several different models based on the architectures and loss functions presented in Sections \ref{subsec:DDPMs} and \ref{subsec:cycleWGAN} above. We choose the best five of those models and present further details about their structure 
in Table \ref{tab:models}. Full details are provided in the Appendix.
\begin{table}[ht]
\centering
\caption{Overview of some of the models trained and their respective architectures.}
\label{tab:models}
\begin{tabular}{llr}
\hline
\textbf{Model Name}  & \textbf{Structure}   \\ \hline 
WGAN-GP-v1 &  Based on loss function in (\ref{loss_cycleWGAN-GP}) and Section \ref{subsec:cycleWGAN}.  \\ \hline 
vDDPM-v1 & The network is a single U-Net. \\ \hline 
DDPM-v2 & Contains U-Net, ResNet and attention block.  \\ \hline 
DDPM-Conv & The ResNet of DDPM-v2 is replaced with ConvNext.   \\ \hline 
DDPM-Aug & Same structure as DDPM-v2, but with data augmentation.   \\ \hline 
\end{tabular}
\end{table}

To train these models a number of different fingerprint datasets were used. The number of images within each of the datasets used varied considerably. Information about the number of images within the datasets for live fingerprint synthesis is provided in Table \ref{tab:dataset}. Similarly, information about the number of images within each of  the datasets used for fingerprint-to-fingerprint transformation is provided in Table \ref{tab:dataset_spoof}. Full details of the model structure and hyperparameters for our CycleWGAN-GP model are found in the Appendix.
\begin{table}[ht]
\centering
\caption{Number of live images within each of the seven datasets used for live fingerprint synthesis.
}
\label{tab:dataset}
\begin{tabular}{llllllll}
\hline
\textbf{Name}  & S1-L & S2-L & S2-L-1 &S2-L-2& S3-L & S4-L & S5-L\\ \hline
\textbf{Amount}  & 8622 & 45043 & 22609 & 22434 & 37138 & 9163 & 9149\\ \hline 
\end{tabular}
\end{table}

\begin{table}[ht!]
\centering
\caption{Number of images within each dataset used for fingerprint-to-fingerprint transformation. 
The spoof images are made of different materials and molds such as for instance gelatine, latex or wood glue.}
\label{tab:dataset_spoof}
\begin{tabular}{llr}
\hline
\textbf{Name}  & \textbf{Conditions} & \textbf{Amount}  \\ \hline 

S6-L  & Live & 22434  \\ \hline 
S6-SM1  & Spoof Material 1 & 3000  \\ \hline 
S6-SM2  & Spoof Material 2 & 3000  \\ \hline 
S6-SM3  & Spoof Material 3 & 3000  \\ \hline 
S7-L  & Live & 14993  \\ \hline 
S7-SM4  & Spoof Material 4 & 3000  \\ \hline 
\end{tabular}
\end{table}

\subsection{Metrics\label{subsec:metrics}}
We employ different metrics in order to help us evaluate 
the generated fingerprints. Both the Fr\'{e}chet Inception Distance (FID) \cite{metric:FID} and the Kernel Information Distance (KID) \cite{metric:KID} 
are used to estimate the dissimilarity between generated and real datasets. 
The FID score in particular has the capability of also evaluating whether the diversity of the generated images mimics that of the real ones \cite{metric:FID}.

Similarly, Precision, Recall, Density and Coverage (PRDC) are deemed to be important metrics to further 
clarify and evaluate the relationship between two distributions. 
The standard interpretation of these metrics however does not apply for our work since they are 
designed for classification tasks and not 
generative tasks. Consequently, these conventional interpretations are not suitable for evaluating the performance of generative networks. 

The classical definitions of precision and recall were extended to distributions in \cite{sajjadi2018} in order to be able to evaluate generative models. Improved estimation procedures of these were presented in \cite{metric:pr}, that we have adopted as metrics in our work.
These metrics have demonstrated their ability to capture essential factors in image generation tasks. 

The precision and recall metrics that we employ also suffer from a few shortcomings. 
 These metrics lack the ability to discern a flawless correspondence between two identical distributions, making it impossible for instance to attain a precision score of 1. Furthermore, their outcomes can be significantly affected by variations in the hyperparameter representing the number of nearest neighbor elements. Lastly, it is worth noting that these metrics may exhibit vulnerability to outliers, as there exists a potential for generated samples to be erroneously encompassed by the manifold created by actual outliers in specific scenarios.

Finally, as an added level for evaluating the generated images we use biometric assessments: False Accept Rate (FAR) for matching, to evaluate uniqueness, and histograms of spoof detection score, to evaluate the amount of distinct spoof characteristics. These are based on proprietary solutions that were developed and extensively validated by Precise Biometrics AB and are currently used in the real world. 
The FAR, also referred to as the False Match Rate (FMR) \cite{metric:FAR}, is used to identify similarities among generated images and between generated images and real images from the training dataset. 
This has the dual purpose of assessing the uniqueness of the images and gauging whether the generated data poses a similar level of challenge for the matching algorithm as the original data.
\section{Results\label{sec:results}}


\subsection{Live Fingerprint Synthesis}
\subsubsection{Generative Modeling Performance Metrics}
We employed a group of metrics, including FID, KID and PRDC (see Section \ref{sec:methods}), to evaluate the quality and 
diversity of the synthetic fingerprints. 
To better understand the fingerprint synthesis evaluation metrics, we first carried out a number of baseline experiments where these metrics were calculated on pairs of real fingerprint datasets. Table \ref{tab:NFS-RL} presents these baseline scores that can be compared to the scores of generated fingerprint datasets paired with their real fingerprint training data, which are shown in Table \ref{tab:NFS-G}. 

We present a comprehensive overview of five distinct generative models utilized to create synthetic fingerprints from the same dataset in Table \ref{tab:NFS-G}. Our approach encompasses the implementation of four diverse DDPM models alongside a WGAN-GP type model.
Specifically, the WGAN-GP-v1 model yielded an FID score of 16.43 and PRDC scores of 0.51, 0.42, 0.37, and 0.32, respectively. Similar outcomes were observed with the vanilla DDPM model, denoted as vDDPM-v1, whose details are provided in Table \ref{tab:vDDPM} in the Appendix. Notably, this version of DDPM involves a single U-Net within the deep generative model.
Furthermore, we conducted assessments on an advanced DDPM variant, denoted as DDPM-v2 which produced the best results from all models compared in that table. This model incorporates a more complex architecture, featuring both a ResNet and an attention block. Comprehensive details regarding this model and its settings can be found in Table \ref{tab:vDDPM}.
\begin{table*}[ht!]
\centering
\caption{Evaluation of the baseline of all the metrics by hypothesizing one of the real datasets is synthetic. 
S2-L-1 and S2-L-2 are two fingerprint datasets collected with almost the same fingers (4 out of 300 fingers are only present in one of the datasets) from two different devices equipped with the same fingerprint sensor, but with manufacturing differences leading to systematic differences in image intensity histograms. Note: S2-L-2-1 and S2-L-2-2 contain 10,000 different fingerprint images separately, both collected from S2-L-2. }
\label{tab:NFS-RL}
\begin{tabular}{llrrrrrrrr}
\hline
\textbf{Real-A}    & \textbf{Real-B}               & \textbf{FID$\downarrow$}   & \textbf{KID$\downarrow$}                & \textbf{Precision$\uparrow$} & \textbf{Recall$\uparrow$} & \textbf{Density$\uparrow$} & \textbf{Coverage$\uparrow$} \\ \hline 



S2-L-1       &S2-L-2        & 5.93 & 0.0015 & 0.57 & 0.61 & 0.43 & 0.42 \\ \hline

S2-L-2-1    &S2-L-2-2   & \textbf{1.40} & \textbf{0.00} & \textbf{0.86} & \textbf{0.86} & \textbf{1.00} & \textbf{0.87} \\ \hline

S2-L       &S1-L           & 69.24 & 0.0633 & 0.24 & 0.11 & 0.12 & 0.06 \\ \hline

S2-L       &S3-L         & 154.68 & 0.1729 & 0.02 & 0.04 & 0.01 & 0.01 \\ \hline

S2-L       &S4-L          & 51.32 & 0.0515 & 0.55 & 0.57 & 0.42 & 0.29 \\ \hline

S2-L       &S5-L         & 46.59 & 0.0499 & 0.56 & 0.59 & 0.45 & 0.31 \\ \hline
\end{tabular}
\end{table*}

\begin{figure*}[ht!]
    \centering
    \includegraphics[width=0.72\textwidth]{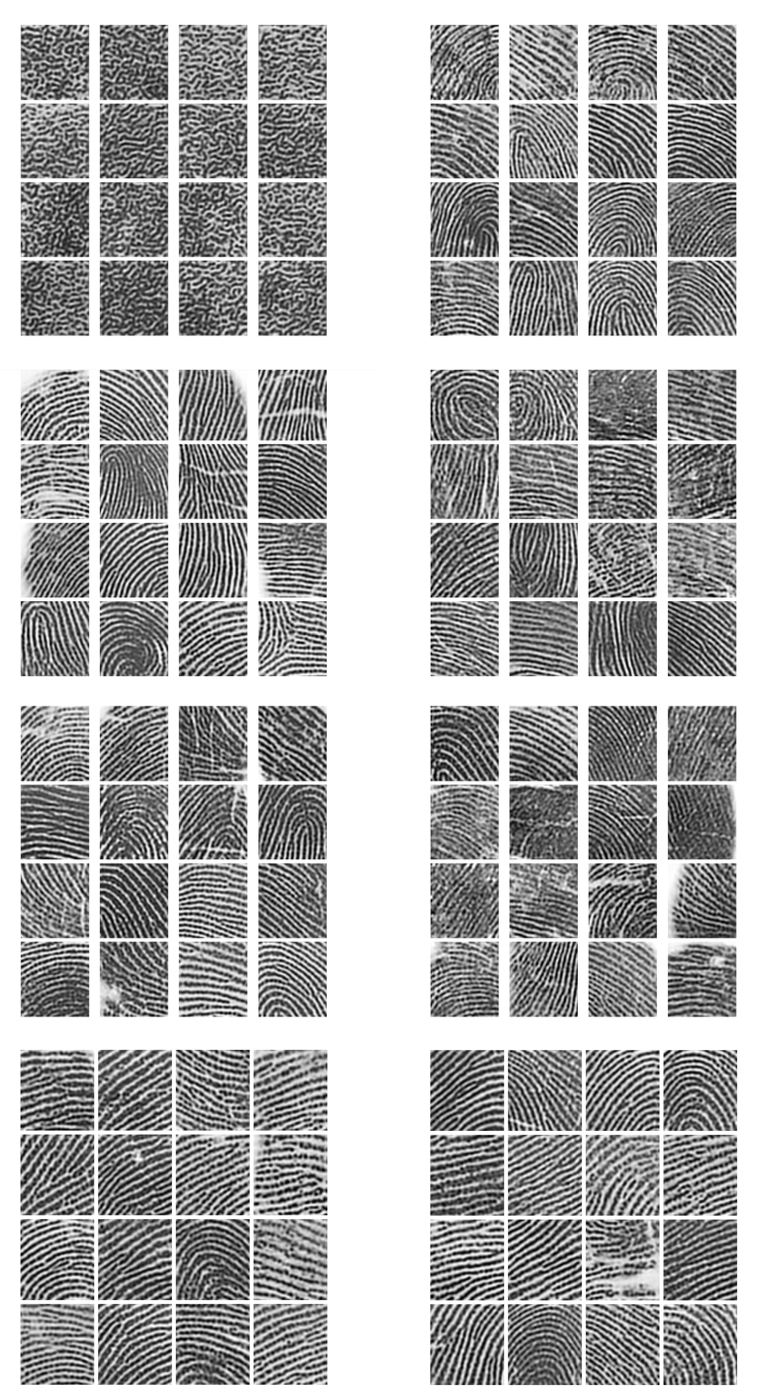}
        
    \caption{Generated live fingerprints at epochs 1, 50, 100, 150, 200, 400, 800 (last row - left) and 1000 (last row -right), from model structure DDPM-v2 trained on the S3-L dataset.
}
        
    \label{fig:gen_samples}
\end{figure*}

\begin{table*}[ht!]
\centering
\caption{Evaluation of synthetic fingerprint images generated by 5 different models trained on one large common dataset S2-L with around 45,000 images in total. 
See Table \ref{tab:models} for information on models. DDPM-v2 seems to outperform all the other models as achieved the best performance across all the metrics. 
}
\label{tab:NFS-G}
\begin{tabular}{lrrrrrrr}
\hline
\textbf{Model name}                    & \textbf{FID$\downarrow$}   & \textbf{KID$\downarrow$}                & \textbf{Precision$\uparrow$} & \textbf{Recall$\uparrow$} & \textbf{Density$\uparrow$} & \textbf{Coverage$\uparrow$} \\ \hline

WGAN-GP-v1
& 16.43 & 0.0136 & 0.51 & 0.42 & 0.37 & 0.32 \\ \hline

vDDPM-v1
& 15.99 & 0.0137 & 0.59 & 0.50 & 0.45 & 0.35 \\ \hline

DDPM-v2
&  \textbf{15.78} &  \textbf{0.0134} &  \textbf{0.69} &  \textbf{0.53} &  \textbf{0.63} &  \textbf{0.41} \\ \hline

DDPM-Conv 
& 42.67 & 0.0497 & 0.65 & 0.56 & 0.60 & 0.34 \\ \hline

DDPM-Aug 
& 21.18 & 0.0211 & 0.53 & 0.60 & 0.39 & 0.29 \\ \hline


\end{tabular}
\end{table*}
To offer a more comprehensive exploration of model-data combinations, we conducted an additional model training phase. This new model, which shares the same architecture as DDPM-v2, was trained on a more secure and openly accessible dataset. On this dataset, our model achieved the following performance metrics: FID$=42.47$, KID$=0.0526$, Precision$=0.81$, Recall$=0.41$, Density$=0.92$, and Coverage$=0.46$. Subsequently, we present the generated fingerprints from this model in Figure \ref{fig:gen_samples} later below.


\subsubsection{Fingerprint Matching Metrics}
As described in Section \ref{sec:methods}, we use scores from a fingerprint matching algorithm obtained from Precise Biometrics as an additional ``external'' measure to evaluate the uniqueness of the synthetic fingerprints, and their difficulty levels when used for matching. For each image pair type (e.g.\ real vs.\ real, real vs.\ DDPM-v2-generated) we compute matching scores for 5 million pairs. 
For real images we distinguish between impostors, i.e.\ images from different persons/fingers, and genuines, i.e.\ images from the same person and finger. If one or both of the two images in a pair is synthetic we call it an ``assumed impostor'' since synthetic images are not based on specific fingers, although by chance they may resemble the same finger.


To gain insights from the score distributions, we calculate False Accept Rates (FARs) for the different types of image pairs. The FAR is a critical metric for assessing the performance of a matching algorithm, and conversely, it gives information about how difficult a certain dataset is for the algorithm. The FAR determines the rate at which the system incorrectly accepts an impostor as a genuine user. It is heavily affected by the tail of the score distribution since typical FARs are very low.

For the analysis, we establish a score threshold that defines when a real/assumed impostor is erroneously accepted as a legitimate user. This threshold is carefully set to achieve a specific baseline FAR level on real impostors. For example, if we target a baseline FAR of 1e-6, it signifies that the matching algorithm will mistakenly accept one out of every million real impostor pairs as genuine. Based on this score threshold we can compute FARs for the synthetic--synthetic and synthetic--real image pairs.

Similar FARs between real impostors and synthetic--synthetic image pairs means that the synthetic dataset presents a similar level of difficulty as the real dataset. Moreover, the FARs of synthetic--real image pairs gives a nuanced view of similarities between the synthetic and the real data. A higher FAR of synthetic--real image pairs than for real impostors means that the synthetic images are less dissimilar to the real images that it was trained on than the typical dissimilarity between fingerprints from different real fingers.

\begin{figure*}[ht!]
    \centering
        \includegraphics[width=.45\textwidth]
        {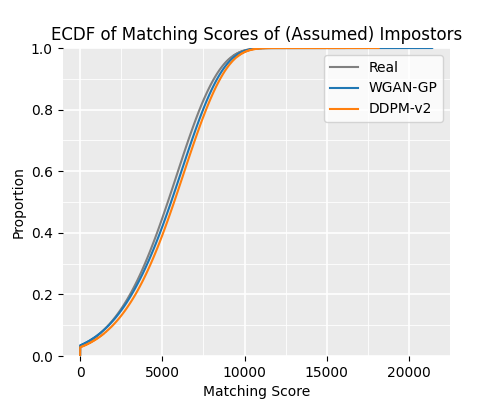}
        \includegraphics[width=.45\textwidth]
        {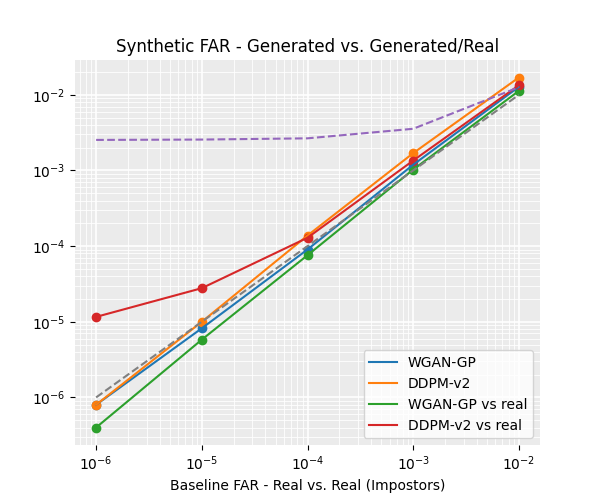}
    \caption{Evaluation based on a matching function applied to pairs of real fingerprint images and/or generated fingerprints from the DDPM-v2 and WGAN-GP-v1 models. 
    Left: Cumulative distribution of the matching scores of real impostors (gray) and assumed impostors (blue and yellow), where the assumed impostors are random pairs of generated images.
    Right: Synthetic FAR for assumed impostors that are either pairs of generated images or a generated image paired with a real image. The threshold for false accepts is set to achieve a baseline FAR on real impostors (x-axis). 
    The synthetic FARs for matching random pairs of generated images are shown in blue and yellow, and the corresponding FARs for matching random generated images with random real images from the training data are shown in red and green.
    The dashed grey line represents identical synthetic and baseline FAR. The purple dashed line represents the FAR that is achieved for random image pairs from the training data, i.e. a mix of genuines (images from same finger) and impostors.
    }
    \label{fig:FAR}
\end{figure*}

In Figure \ref{fig:FAR} (left), we illustrate the cumulative distribution of matching scores for three distinct datasets: one comprising real impostor pairs from the training data and the two other composed of randomly paired generated images, which we consider as assumed impostors. A higher matcher score signifies that an assumed impostor pair is more likely to be a genuine match. The results are shown separately for the WGAN-GP-v1 and vDDPM-v2 models, providing insights into how difficult the generated datasets are for the matching algorithm as compared to real impostors. 

On the right side of Figure \ref{fig:FAR}, we present a detailed analysis focusing on synthetic–synthetic image pairs and synthetic–real image pairs. The key observation involves the matcher's assignment of scores to these image pairs in comparison to real impostor pairs. When the synthetic–synthetic impostor pairs receive higher scores than real impostor pairs, the result is an elevation of the synthetic FAR beyond the baseline level, represented by the dashed grey line on the right side of Figure \ref{fig:FAR}. The dashed grey line indicates an equivalent synthetic FAR compared to the baseline. As a contrast, the purple dashed line in the same figure illustrates random pairings from the training data, encompassing both genuine and impostor pairs. It's worth noting that if the generative models had merely duplicated the training data without generating unique images, the FARs for synthetic–real image pairs would align with the purple dashed line. Conversely, if all generated images were as unique as genuine fingerprints from distinct fingers, the FARs would align with the grey dashed line. Figure \ref{fig:FAR} therefore provides further insights into the extent to which the models draw inspiration from the training data in terms of matching fingerprint patterns.

\subsection{Fingerprint-to-Fingerprint Transformation}\label{sec:FFT}

We 
trained a CycleWGAN-GP network to transform a live fingerprint to a spoof fingerprint. 
Furthermore, we confined the spoof fingerprint images to represent spoofs based on specific physical materials such as for instance gelatine, latex or wood glue \cite{spoof-handbook}. We consider three different types of spoof materials in this paper. 
To train our neural network we use a dataset of 22,000 live fingerprints and a spoof fingerprint dataset containing 3000 spoof fingerprints from each 
material. 

We compare Real spoof (RS) with Gen spoof (GS) fingerprints in Table \ref{tab:table-cycleWGAN-280223-01-RL}. In that same table, we provide an assessment of the similarity between Real live (RL) and Real spoof (RS) as well as Real live (RL) and Gen spoof (GS) fingerprints.

\begin{table}[ht!]\footnotesize
\centering
\caption{Evaluation of fingerprint-to-fingerprint transformations between \textit{Real live (RL)}, \textit{Real spoof (RS)} and \textit{Gen spoof (GS)}. The original datasets are \textit{Real live (RL)} and \textit{Real spoof (RS)}. The generated dataset is \textit{Gen spoof (GS)}. The up arrows indicate larger values are better, and vice versa. The model was trained with a sample size of 1000 live and spoof images respectively. 
}
\label{tab:table-cycleWGAN-280223-01-RL}
\resizebox{\textwidth}{!}{%
\begin{tabular}{llllrrrrrrr}
\hline
\textbf{Material}  &  \textbf{Dataset-A}   & \textbf{Amount} & \textbf{Dataset-B} & \textbf{Amount}   & \textbf{FID$\downarrow$}  & \textbf{Pre.$\uparrow$} & \textbf{Rec.$\uparrow$} & \textbf{Den.$\uparrow$} & \textbf{Cov.$\uparrow$} \\ \hline 

1 & Real live (RL) & 22434 & Real spoof (RS)   & 3150   & 67.46  & 0.37  & 0.58 & 0.21  & 0.19  \\ \hline
1 & Real live (RL) & 22434 & Gen spoof (GS)    & 22434  & 64.61  & 0.34  & 0.60 & 0.22  & 0.20  \\ \hline
1 & Real spoof (RS)  & 3150  & Gen spoof (GS)    & 22434  & 47.28  & 0.75  & 0.46 & 0.84  & 0.53  \\ \hline

\hline

2 & Real live (RL) & 22434 & Real spoof (RS)   & 3065   & 103.86 & 0.44 & 0.33 & 0.30 & 0.09 \\ \hline
2 & Real live (RL) & 22434 & Gen spoof (GS)    & 22434  & 106.80 & 0.30 & 0.56 & 0.22 & 0.18 \\ \hline
2 & Real spoof (RS)  & 3065  & Gen spoof (GS)    & 22434  & 23.34  & 0.58 & 0.53 & 0.52 & 0.51 \\ \hline

\hline

3 & Real live (RL) & 22434 & Real spoof (RS)   & 2993   & 162.70 & 0.52 & 0.39 & 0.45 & 0.05 \\ \hline
3 & Real live (RL) & 22434 & Gen spoof (GS)    & 22434  & 167.54 & 0.56 & 0.34 & 0.47 & 0.07 \\ \hline
3 & Real spoof (RS)  & 2993  & Gen spoof (GS)    & 22434  & 16.67  & 0.66 & 0.53 & 0.73 & 0.57  \\ \hline

\end{tabular}%
}
\end{table}

\begin{figure*}[ht!]
    \centering
        \includegraphics[width=1\textwidth]{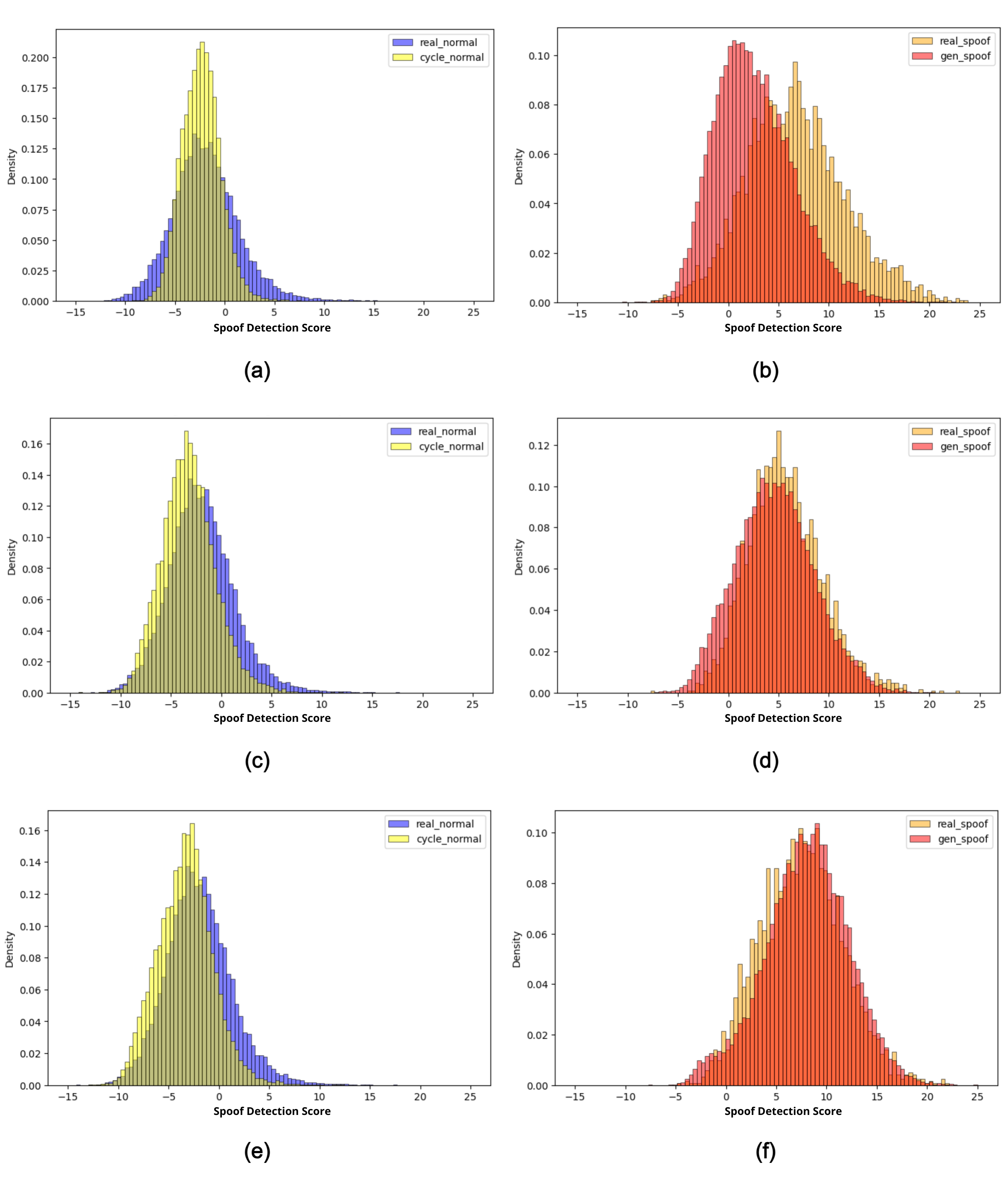}
    \caption{The spoof detection scores between Real live (RL) and Cycle live (CL), Real spoof (RS) and Generated spoof (GS). The subfigures (a) and (b) come from material 1, (c) and (d) come from material 2, and so on. All left-side figures illustrate the similarity between RL and CL to verify the consistency of CycleWGAN-GP. All right-side figures illustrate the similarity between RS and GS.}
    \label{fig:FFT}
\end{figure*}

We applied a spoof detection function from Precise Biometrics to further evaluate the generated spoof fingerprints. The results are shown in Figure \ref{fig:FFT}, and a lower spoof detection score indicates that the fingerprint is less likely to be a spoof, and vice versa.


\clearpage
\section{Discussion\label{sec:discussion}}


We used generative adversarial networks and diffusion models to synthesize high-quality live and spoof fingerprint images while preserving their uniqueness and complexity. Our techniques have the capability to generate random fingerprints from pure noise and incorporate style transfer methods, enabling the fusion of global finger structures with local features from diverse fingerprints. To comprehensively evaluate our proposed approach, we employ established metrics such as the Fréchet Inception Distance (FID) and the False Acceptance Rate (FAR). These metrics provide a rigorous assessment of the generated fingerprints in terms of their quality, diversity, and resemblance to real fingerprints.

\subsection{Live Fingerprint Synthesis}

In our exploration of deep generative models for live fingerprint synthesis, we aim to not only highlight their performance but also delve into previously unaddressed training and image generation intricacies. We present a comprehensive analysis of our experiments.

Initially, we began with the widely recognized DCGAN architecture due to its established efficacy. However, it became apparent that DCGAN, with its relatively simple configuration, underperformed on our dataset. This could be attributed to limited tuning and network depth exploration. Therefore, we shifted our focus to the more robust WGAN-GP architecture, which 
clearly outperformed DCGAN in generating realistic images. It's worth noting that WGAN-GP often converged within just 50 epochs, but generating high-quality fingerprints sometimes required up to 500 epochs, presenting potential optimization challenges.

Our research also extended to diffusion models, where we implemented various types with similar structures. 
Diffusion models have higher computational cost than WGAN-GP, but they showcased superior synthesis quality and avoided issues like model collapse.

To provide a baseline for evaluating synthetic fingerprints, we analyzed the metrics presented in Table \ref{tab:NFS-RL}, which stem from comparisons among real fingerprint datasets. These results serve as a benchmark for assessing the quality of generated datasets. 
Notably, datasets S2-L-2-1 and S2-L-2-2 in Table 4 exhibited remarkable similarity, with minimal FID scores, KID scores of 0, and significantly higher PRDC compared to other pairs. These images were collected under the same conditions, with the same device, and with the same fingers. Conversely, datasets S2-L-1 and S2-L-2, which also had the same fingers (4 out of 300 fingers were only in one of the datasets), but with significant differences in intensity histograms due to that they were collected on different devices. These datasets had a FID score of 5.93 and precision, recall, density, and coverage in the range 0.42--0.61. The remaining pairs involved scores calculated across real datasets from different sensors, and with different fingers.

Table \ref{tab:NFS-G} showcases the results of live fingerprint synthesis using diverse generative models. WGAN-GP achieved a FID of 16.43, indicating a close resemblance to real fingerprints. Remarkably, DDPM-v2 outperformed all others with the lowest FID of 15.78 and top scores across all metrics, signifying superior similarity. Additionally, we computed FAR for evaluation, and visual comparisons are available in Figure \ref{fig:FAR}.

In Figure \ref{fig:FAR} (left), we observe that the cumulative distribution of matcher scores for generated image pairs closely resembles that of real impostors from the training data, albeit slightly shifted to the right. This suggests that generated pairs are slightly more similar to each other than real impostors. However, for high-security applications like fingerprint recognition, the right tail of the distribution becomes crucial. Thus, we computed synthetic FAR at different security levels, as shown in Figure \ref{fig:FAR} (right).

For DDPM-v2 paired with real images, the synthetic FAR exceeds the baseline FAR for high-security levels (e.g., FAR 1e-5 and 1e-6), indicating similarities found by the matcher between generated and training data. Notably, the FAR remains well below what would be expected if the generated data were mere copies of the training data. Visual inspections revealed distinct differences even in images with the highest matching scores.

Conversely, when images from WGAN-GP are paired with real images, we observe lower FAR than the baseline, indicating clear distinctions between generated and training data. This matching function provides valuable insights into how generative models operate and the degree of similarity between generated and training images.

\subsection{Fingerprint-to-Fingerprint Transformation}

The high-quality synthetic live fingerprints we obtained encouraged us to expand the research focus towards generating physical spoof fingerprints 
made out of different materials such as gelatine, latex or wood glue \cite{spoof-handbook}. Since the volume of spoof fingerprints is typically limited, directly using a spoof fingerprint dataset to train a generative model faces challenges in terms of feasibility and efficacy. Instead, we aimed to leverage the vast advantage of large datasets of live fingerprints and employed style transfer techniques to generate spoof fingerprints.

In Table \ref{tab:table-cycleWGAN-280223-01-RL}, we list the results of three spoof fingerprint materials corresponding to three different cycleWGAN-GP models. As we can see in the table, when comparing real spoof (RS) with generated spoof (GS) fingerprints, Material 1 obtains the highest FID of 47.28. 
This indicates that generated data based on Material 1 is the least similar to its real counterpart compared to those from Material 2 which has an FID of 23.34 or Material 3 which has an FID of 16.67. 
These results are reflected in Figure \ref{fig:FFT} (b), (d) and (f) respectively, where more overlapped histograms indicate better similarity and lower FID score. Another way to evaluate the quality of GS is by assessing the similarity between real live (RL) and real spoof (RS) and the similarity between real live (RL) and generated spoof (GS). 
Specifically, we find that for Materials 1,2 and 3, the FID scores of (RL-RS, RL-GS) are respectively, for each material, (67.46, 64.61), (103.86, 106.80) and (162.70, 167.54). 
Notably, cycleWGAN-GP exhibited unexpected behavior in cycle-generated fingerprints since it produced cycle images that were on average more distinct from spoof images than their real counterparts, evident in Fig.\ \ref{fig:FFT} (c) and (e). This requires further evaluation to discern underlying changes.

In summary, unlike generating fingerprints from noise, fingerprint-to-fingerprint transformation requires a smaller amount of data to generate high-quality synthetic fingerprints paired with existing fingerprints in other conditions.
Transformation quality correlated with spoofiness (i.e.\ how easy it is to detect it as a spoof); higher spoofiness yielded better transformations. Further exploration is however needed to better understand cycleWGAN-GP's effects on fingerprint characteristics.

\section{Conclusions\label{sec:conslusions}}

In this study, we propose a methodology for synthesizing high-quality patch size fingerprint images while preserving their unique and complex features. 
To achieve this, we leverage the powers of a generative adversarial network (WGAN-GP) and a Denoising Diffusion Probabilistic Model (DDPM), which, to the best of our knowledge, has not been done previously. 
Additionally, we demonstrate the use of style transfer techniques through a cycle autoencoder network (cycleWGAN-GP), enabling the seamless integration of global fingerprint structures with localized features from diverse fingerprint patches.
To assess the quality of the synthesized results, we employ several evaluation metrics including Fréchet Inception Distance (FID), Kernel Inception Distance (KID), Precision, Recall, Density, and Coverage (PRDC).

For evaluating the quality of the generated fingerprints, we recommend employing Fréchet Inception Distance (FID) for assessing similarity and FAR for assessing uniqueness. Using Inception Score (IS) is generally discouraged since it focuses on diversity over quality and can also be influenced by background noise or other irrelevant details thus lacking discrimination. Precision, Recall, Density, and Coverage (PRDC) on the other hand can be used to provide additional insights for a comprehensive understanding of the results.

Our investigation shows that DDPM outperforms other models in generating the most similar and high-quality fingerprints, but at the cost of losing some of the creativity and uniqueness. On the contrary, while the quality of the fingerprints generated by WGAN-GP is slightly lower, they outperform those generated by DDPM on the uniqueness assessment. Furthermore, cycleWGAN-GP can generate high-quality spoof fingerprints in multiple conditions given fewer training data, while maintaining a reasonable computational cost and network consistency.



Our research has also revealed a significant connection between the success of fingerprint transformation techniques and 
the detectability of spoof fingerprints, a characteristic we described previously as spoofiness.
When a fingerprint dataset exhibits higher levels of spoofiness, it leads to greater success in applying these transformations techniques in both of our models.

Considering the challenges associated with obtaining publicly available, high-quality fingerprint datasets, our work contributes to the development of a GDPR-compliant public dataset that encompasses unique fingerprints. As part of our future research endeavors, we propose constructing deep generative models tailored to fingerprints under various conditions, such as spoof, wet, cold, and others. 

\section*{Acknowledgements}
The work for A.~S. and D.~L. is partially supported by grants from eSSENCE no. 138227, Vinnova no. 2020-033375 and Rymdstyrelsen no. 2022-00282. Training and evaluation of models were run on GPU servers provided by Precise Biometrics. We would also like to thank Ulf Holmstedt, Johan Windmark and Anders Olsson at Precise Biometrics for helpful discussions.

\section*{Appendix A - Specific Network Architectures\label{sec:appendixA}}

The layers, hyperparameters and specific configurations used in each of our base networks are provided here for full transparency.

\subsection*{DDPM Architectures}

We initially implemented a vanilla DDPM version (or vDDPM-v1), in which the deep learning model only consists of a typical U-Net structure within a type of layer normalization and a sigmoid linear unit (SiLU) activation function. This is a typical U-Net that can first down-sample the input image from 112x112 to 3x3, and then up-sample it back to 112x112. The output image is the estimation of the noise at that time step.

Multiple setups of kernel size and the number of channels have been tried when training this vDDPM-v1 model. The main settings for this model are presented in  Table \ref{tab:vDDPM}.
\begin{table}[ht]
\centering
\caption{Setting for models vDDPM-v1 and DDPM-v2.}
\label{tab:vDDPM}
\begin{tabular}{lrr}
\hline
\textbf{Model Name}  & vDDPM-v1 & DDPM-v2\\ \hline
\textbf{Image Size}  & Padded to 112x112 & Padded to 112x112\\ \hline
\textbf{Time Embedding}  &  Sinusoidal Embedding& Sinusoidal Embedding\\ \hline
\textbf{Time Schedule}  & Linear Schedule & Cosine Schedule \\ \hline
\textbf{Time Steps}  &  1000 & 1000\\ \hline
\textbf{Noise Level}  &  [1e-5, 1e-2] &   [1e-5, 1e-2]\\\hline
\textbf{Batch Size}  &  64 & 64\\ \hline
\textbf{Learning Rate}  &  1e-4 & 1e-4\\ \hline
\textbf{Epochs}  &  500 & 500\\ \hline
\textbf{Loss}  &  MSE & Huber\\ \hline
\textbf{Optimizer}  &  Adam & Adam\\ \hline
\end{tabular}
\end{table}


After achieving good fitting results with  vDDPM-v1, we decided to increase the complexity of the model. We therefore introduced various modules such as attention and ResNet. We call this model DDPM-v2 and provide its details in the same Table \ref{tab:vDDPM}.


\subsection*{CycleWGAN-GP Architecture}

We adopt the base architecture for our generative networks from Jun-Yan Zhu et al. \cite{cycleGAN} who have shown impressive results for style transfer, object transfiguration, season transfer, photo generation from paintings and photo enhancement.

The generator network inputs training images padded to $128 \times 128$. This network contains three convolutions, six residual blocks \cite{ResNet}, two fractionally-strided convolutions with stride $\frac{1}{2}$, and one convolution that maps features to grayscale. Similar as Jun-Yan Zhu et al. \cite{cycleGAN}, we use instance normalization \cite{InsNorm}. Below, we follow the naming convention used in the Johnson et al.’s Github repository.
    
Let $c7s1-k$ denote a $7 \times 7$ Convolution-InstanceNorm-ReLU layer with $k$ filters and stride 1. The notation $dk$ denotes a $3 \times 3$ Convolution-InstanceNorm-ReLU layer with $k$ filters and stride 2. Reflection padding was used to reduce artifacts. $Rk$ denotes a residual block that contains two $3 \times 3$ convolutional layers with the same number of filters on both layer. $uk$ denotes a $3 \times 3$ fractional-strided-Convolution-InstanceNorm-ReLU layer with $k$ filters and stride 12. The network with 6 residual blocks consists of: 
$$c7s1-64,d128,d256,R256,R256,R256,R256,R256,R256,u128,u64,c7s1-3.$$
The discriminator networks contains four convolutions, which aim to classify whether the image is real or fake.
    
Let $Ck$ denote a $4 \times 4$ Convolution-InstanceNorm-LeakyReLU layer with $k$ filters and stride 2. After the last layer, we apply a convolution to produce a 1-dimensional output. We do not use InstanceNorm for the first C64 layer. We use leaky ReLUs with a slope of 0.2. The discriminator architecture is: $C64-C128-C256-C512$.

We follow the same guidelines as Jun-Yan Zhu et al. \cite{cycleGAN}. 
We use the Adam solver [26] with a batch size of 1. All networks were trained from scratch with a learning rate of 0.0002. We keep the same learning rate for the first 100 epochs and linearly decay the rate to zero over the next 100 epochs. The remaining details and hyperparameters for this CycleWGAN-GP architecture can be found in Table \ref{tab:cycleGANparameters}.
    \begin{table}[ht]
    \centering
    \caption{Model structure of CycleWGAN-GP model.}
    \label{tab:cycleGANparameters}
    \begin{tabular}{lr}
    \hline
    \textbf{Model Name}  & CycleWGAN-GP \\ \hline
    \textbf{Input}  &  Padded to 128x128\\ \hline
    {\boldmath$\lambda_\textbf{cycle}$}  &  10.0\\ \hline
    {\boldmath$\lambda_\textbf{identity}$}  & 0.5 \\\hline
    {\boldmath$\lambda_\textbf{GP}$}    &   10.0 \\ \hline
    \textbf{Batch Size}  &  1\\ \hline
    \textbf{Epochs}  &  500\\ \hline
    \textbf{Loss}  &  Wasserstein\\ \hline
    \textbf{Optimizer}  &  Adam\\ \hline
    \textbf{Learning Rate}  &  2e-4\\ \hline
    {\boldmath$\beta_1$}  &  0.5\\ \hline
    \end{tabular}
    \end{table}

\bibliographystyle{hindawi_bib_style}

\end{document}